\documentclass[pdflatex,sn-mathphys-num]{sn-jnl}

\usepackage{graphicx}%
\usepackage{multirow}%
\usepackage{amsmath,amssymb,amsfonts}%
\usepackage{amsthm}%
\usepackage{mathrsfs}%
\usepackage[title]{appendix}%
\usepackage{xcolor}%
\usepackage{textcomp}%
\usepackage{manyfoot}%
\usepackage{booktabs}%
\usepackage{algorithm}%
\usepackage{algorithmicx}%
\usepackage{algpseudocode}%
\usepackage{listings}%

\theoremstyle{thmstyleone}%
\theoremstyle{thmstyletwo}%

\theoremstyle{thmstylethree}%

\begin{document}

\title[Article Title]{BarlowTwins-CXR : Enhancing  Chest  X-Ray abnormality localization in heterogeneous data with cross-domain self-supervised learning}


\author{\fnm{Haoyue} \sur{Sheng}}\email{haoyue.sheng@umontreal.ca}

\author{\fnm{Linrui} \sur{Ma}}\email{linrui.ma@umontreal.ca}

\author{\fnm{Jean-François} \sur{Samson}}\email{jean-francois.samson.ccsmtl@ssss.gouv.qc.ca}

\author{\fnm{Dianbo} \sur{Liu}}\email{dianbo@nus.edu.sg}

\abstract{
\textbf{Background:} 
Chest X-ray imaging based abnormality localization, essential in diagnosing various diseases, faces significant clinical challenges due to complex interpretations and the growing workload of radiologists. While recent advances in deep learning offer promising solutions, there is still a critical issue of domain inconsistency in cross-domain transfer learning, which hampers the efficiency and accuracy of diagnostic processes.


This study aims to address the domain inconsistency problem and improve autonomic abnormality localization performance of heterogeneous chest X-ray image analysis, particularly in detecting abnormalities, by developing a self-supervised learning strategy called “BarlwoTwins-CXR”.

\textbf{Methods:} We utilized two publicly available datasets: the NIH Chest X-ray Dataset and the VinDr-CXR. The BarlowTwins-CXR approach was conducted in a two-stage training process. Initially, self-supervised pre-training was performed using an adjusted Barlow Twins algorithm on the NIH dataset with a Resnet50 backbone pre-trained on ImageNet. This was followed by supervised fine-tuning on the VinDr-CXR dataset using Faster R-CNN with Feature Pyramid Network (FPN). The study employed mean Average Precision (mAP) at an Intersection over Union (IoU) of 50\% and Area Under the Curve (AUC) for performance evaluation.

\textbf{Results:} Our experiments showed a significant improvement in model performance with BarlowTwins-CXR. The approach achieved a 3\% increase in mAP50 accuracy compared to traditional ImageNet pre-trained models. In addition, the Ablation CAM method revealed enhanced precision in localizing chest abnormalities. The study involved 112,120 images from the NIH dataset and 18,000 images from the VinDr-CXR dataset, indicating robust training and testing samples.

\textbf{Conclusion:} 
BarlowTwins-CXR significantly enhances the efficiency and accuracy of chest X-ray image-based abnormality localization, outperforming traditional transfer learning methods and effectively overcoming domain inconsistency in cross-domain scenarios. Our experiment results demonstrate the potential of using self-supervised learning to improve the generalizability of models in medical settings with limited amounts of heterogeneous data. This approach can be instrumental in aiding radiologists, particularly in high-workload environments, offering a promising direction for future AI-driven healthcare solutions.
}

\keywords{medical image analysis; chest x-ray; abnormality localization; deep learning; object detection; self-supervised learning; transfer learning; heat map; area under curve; mean Average Precision.}



\maketitle

\section{Introduction}\label{sec1}



Chest X-ray(CXR) is a fundamental and widespread medical diagnostic tool for diagnosing chest diseases. It is efficient and cost-effective, suitable for preliminary screening and diagnosis \cite{ref1}. During the 2019 coronavirus pandemic, CXR was widely used for triaging patients and prioritizing the care order due to its convenience and flexibility. Effective mitigation addresses the lack of availability of computed tomography and reduces the risk of transmission in the room with the CT scanner \cite{ref2}. However, its complex interpretation often requires a highly qualified radiologist to make an accurate diagnosis \cite{ref1}. As the demand for healthcare increases, the workload of radiologists has significantly increased \cite{ref3}. It results in less time to analyze each radiographic image, potentially increasing the risk of diagnostic error. In many areas, especially in developing and remote areas, qualified radiologists are insufficient to cope with the increased demand for healthcare. For instance, Europe has 13 radiologists per 100,000 people, while the United Kingdom has 8.5, and Malaysia has approximately 30 per million population \cite{ref4}. This situation necessitates urgently developing and introducing automated technologies like AI-based image analysis tools to aid radiologists in quicker and more precise CXR image analysis. It will improve the quality of diagnosis and help reduce the workload of doctors.

In recent years, deep learning models have rapidly advanced in various medical image analysis fields of CXR, demonstrating diagnostic accuracy comparable to human experts \cite{ref5}. Object detection plays a more critical role in medical image analysis because it can identify and precisely locate the types of anomalies in the images, providing doctors with more specific and valuable information. However, training these models requires a large amount of annotated data. These annotations must be performed by experienced radiologists for CXR images, as well as for most medical images, making such annotated data not only costly, but also rare, with only a very limited number of public datasets including bounding box information. Although transfer learning is widely regarded as an effective method to solve the problem of scarce labelling data, its application in medical image analysis still faces limitations. This is mainly due to the significant difference in feature distribution between large datasets(such as ImageNet) used for pre-training models and medical imaging datasets. This cross-domain transfer learning disparity that directly applying these pre-trained weights to medical image analysis might not yield the best outcomes, particularly for specialized medical diagnostic applications \cite{ref6}\cite{ref7}.



In order to develop an efficient chest x-ray image analysis method while addressing the problem of data scarcity and domain inconsistency in transfer learning, our study proposed the BarlowTwins-CXR method, and its main contributions are as follows:

\begin{enumerate}
    \item Proposed a new self-supervised two-phase training strategy to diagnose and locate abnormalities in CXR. Bringing self-supervised learning into chest X-ray anomaly localization solves the problem of cross-domain transfer learning differences.
    \item In the first learning phase, our approach leverages self-supervised pre-training with the Barlow Twins algorithm \cite{ref8}, applied to CXR images without annotation. This strategy addresses the challenge of data heterogeneity between pre-trained ImageNet \cite{ref9} models and CXR images. In the second phase, transfer learning on the VinDr-CXR \cite{ref10} dataset is applied to fine-tune the model.

    \item In the experiment employing ResNet50 \cite{ref11} as the backbone architecture, we observed that implementing the BarlowTwins-CXR strategy significantly improved model performance. We observed a 3\% increase in model accuracy on the mean Average Precision benchmark, surpassing the results achieved by directly performing transfer learning from ImageNet pre-trained weights.

    \item Our experiment results demonstrate the potential of self-supervised learning in medical image analysis, especially in the absence of annotated data, paving the way for more efficient and accurate diagnostic aids methods in the future.
\end{enumerate}

The paper is organized as follows: 
First, the related work section presents an overview of the research related to deep learning, self-supervised learning and transfer learning in the field of medical images. The methods section describes the dual-phase training process with dataset setup and preprocessing. The results section shows and compares the results obtained with different image sizes and backbone weights. We also used Heatmap visualization and Linear Evaluation Protocol to highlight the scheme's effectiveness in this section. Finally, the conclusions of this study and suggestions for future research are discussed in the discussion section.

\section{Related Work}\label{sec2}

In recent years, deep learning techniques have excelled in the field of medical imaging, particularly in analyzing CXR images. For example, in terms of disease classification, ChexNet proposed by Pranav Rajpurkar et al. \cite{ref12} outperformed radiologists in detecting chest diseases, when benchmarked against the F1 score.  Neural network models trained with vast amounts of labelled data are capable of identifying features of various pulmonary diseases. In anomaly detection tasks, Sun K X et al. used the YOLOv7 object detection framework to effectively identify and locate lesions in CXR images \cite{ref13}. This achievement is attributed to the advanced image recognition and feature extraction capabilities of neural networks. Additionally, the modified U-net architecture which incorporates attention mechanisms, as proposed by Gusztáv Gaál et al. \cite{ref14}, has made significant strides in accurately segmenting lung structures, thus aiding in detailed analysis and diagnosis of diseases.

Self-supervised learning has recently gained popularity in the field of medical imaging \cite{ref15} and provides an efficient method for utilizing unlabeled data. Initially proposed by Bengio et al., this approach allows models to learn from unlabeled data and extract useful feature representations by training deep networks on unsupervised data \cite{ref16}. Such learning strategy promotes models to capture the intrinsic structure and relationships in data by designing innovative pretext tasks, such as image reconstruction (e.g., Context encoder \cite{ref17}), contrastive learning (e.g., SimCLR \cite{ref18}), or prediction tasks (e.g., rotation prediction \cite{ref19}). In the field of medical imaging, Shekoofeh Azizi et al. used large-scale images for self-supervised learning to improve accuracy and convergence speed significantly in downstream tasks, achieving better performance than models pre-trained on ImageNet \cite{ref20}. Sowrirajan H et al. proposed a pre-trained model based on Momentum Contrast to enhance the representativeness and portability of CXR models \cite{ref21}.

In terms of transfer learning, applying models trained in one domain to another has led to notable success in medical image analysis. Research indicates that well-processed transfer results from ImageNet can improve model performance in the medical imaging domain \cite{ref22}. However, studies by Christos Matsoukas et al. have shown that due to the significant difference in feature distribution between medical and natural images, features learned from natural images may not always be broadly applicable to medical images \cite{ref23}. Various cross-domain adaptive transfer learning methods have been developed to address these challenges, such as unsupervised and semi-supervised learning and sequential domain adaptation techniques. By tuning model parameters, these methods can be better adapted to the characteristics of medical images, improving the performance and accuracy of models in medical image analysis \cite{ref22}.

\section{Methods}\label{sec3}

\subsection{Dataset Selection}\label{subsec3}

\begin{figure}[h]
\centering
\includegraphics[width=0.8\textwidth]{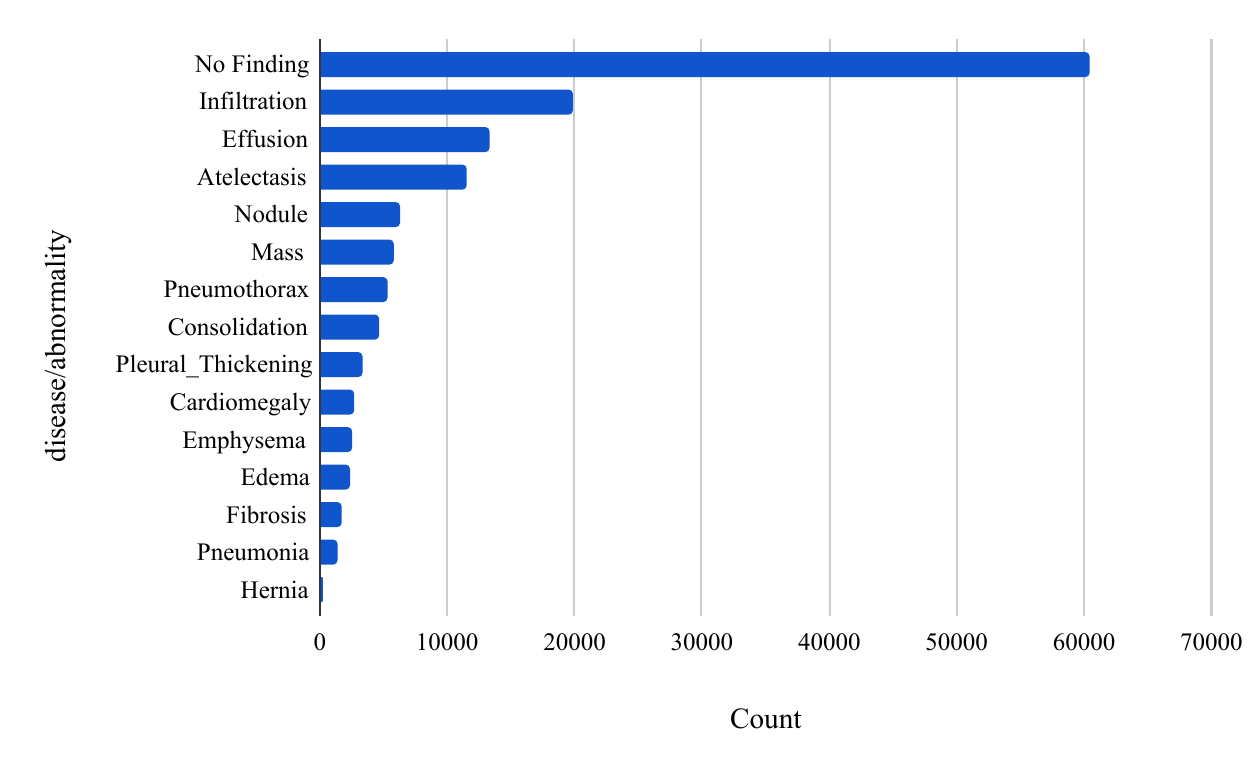}
\caption{Image-level label distribution of the NIH-CXR dataset. }\label{fig1}
\end{figure}

This study utilized two publicly available chest X-ray datasets: the NIH-CXR\cite{ref24} dataset and the VinDr-CXR dataset. The NIH dataset comprises 112,120 posterior-anterior (PA) or anterior-posterior (AP) CXR images from 30,805 patients, covering 14 diseases with image-level annotations, including disease location annotations in some images. The distribution of the NIH-CXR dataset is illustrated in Figure \ref{fig1}.

Meanwhile, the VinDr-CXR dataset is the largest publicly available dataset for adult CXR object detection, which includes 18,000 PA CXR scans. These scans encompass 14 diseases with detailed instance-level bounding box annotations, making it ideal for the fine-tuning phase. 

The VinDr-CXR dataset exhibits a distinct labelling process for its test and training sets. The training set, consisting of 15,000 images, was annotated independently by three radiologists per image. In contrast, the test set, comprising 3,000 images, underwent a more rigorous annotation process. Initially, each image was independently annotated by three radiologists. This is followed by a secondary review phase where these initial annotations are reviewed by two other more experienced radiologists, they communicated with each other to resolve any disagreements and reach a consensus on the final labelling. This meticulous process for the test set created a potential disparity in data distribution compared to the training set. To eliminate any bias it might introduce in our study, we resplit the original training set into new training, validation, and test sets for our experiments.

\begin{figure}[h]
\centering
\includegraphics[width=0.8\textwidth]{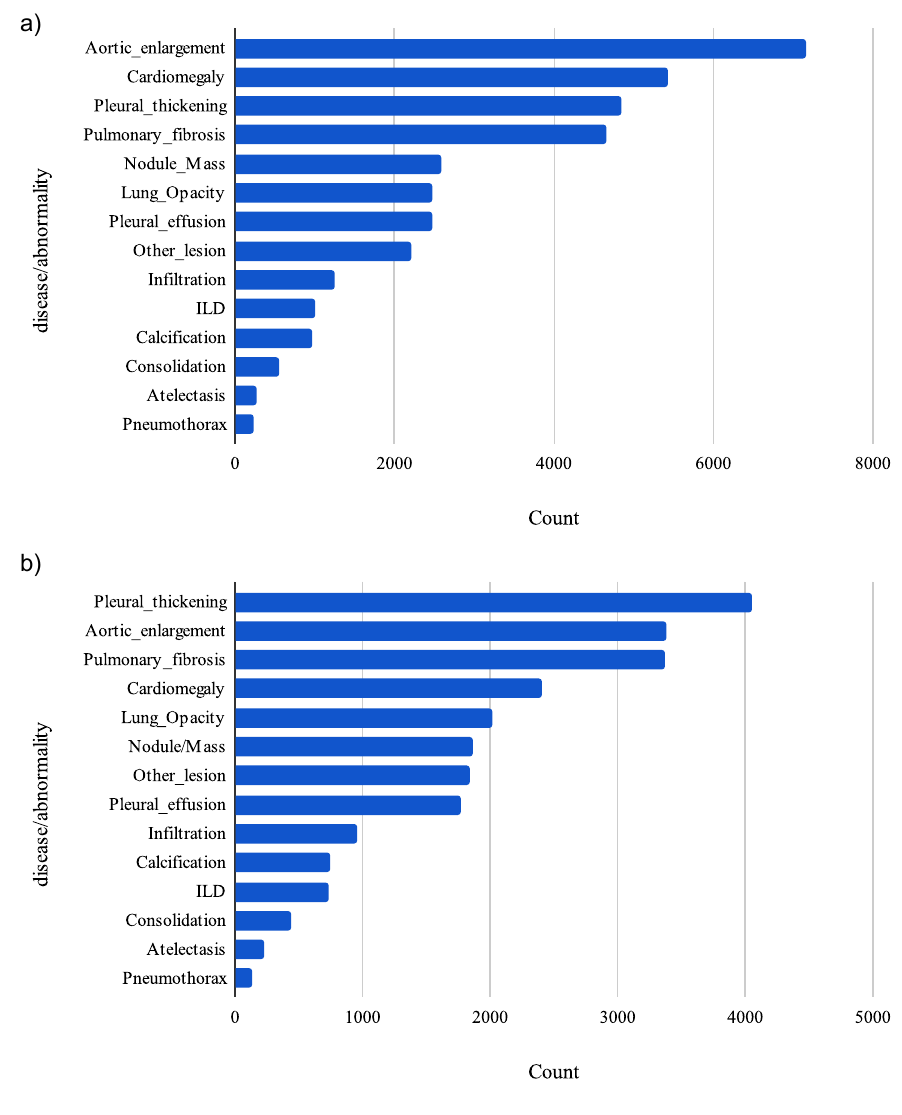}
\caption{Instance-level annotation distribution of VinDr-CXR dataset before(a) and after(b) WBF preprocessing. }\label{fig2}
\end{figure}

To improve the quality of the training data, a Weighted Box Fusion (WBF) \cite{ref25} preprocessing technique was applied to the VinDr-CXR training set. The WBF involves calculating the weighted average of each set of duplicate bounding boxes to create a single fused bounding box. Such a preprocessing step is crucial for reducing annotation redundancy and improving target area representation in the dataset. Figure \ref{fig2} shows the data distribution of VinDr-CXR before and after WBF preprocessing.

We chose the VinDr-CXR dataset not only because it is the largest publicly available dataset for adult CXR object detection, but also because of the high level of diversity and richness of its data.

\subsection{Dual-Phase Training Process}\label{subsec3}

\begin{figure}[h]
\centering
\includegraphics[width=0.8\textwidth]{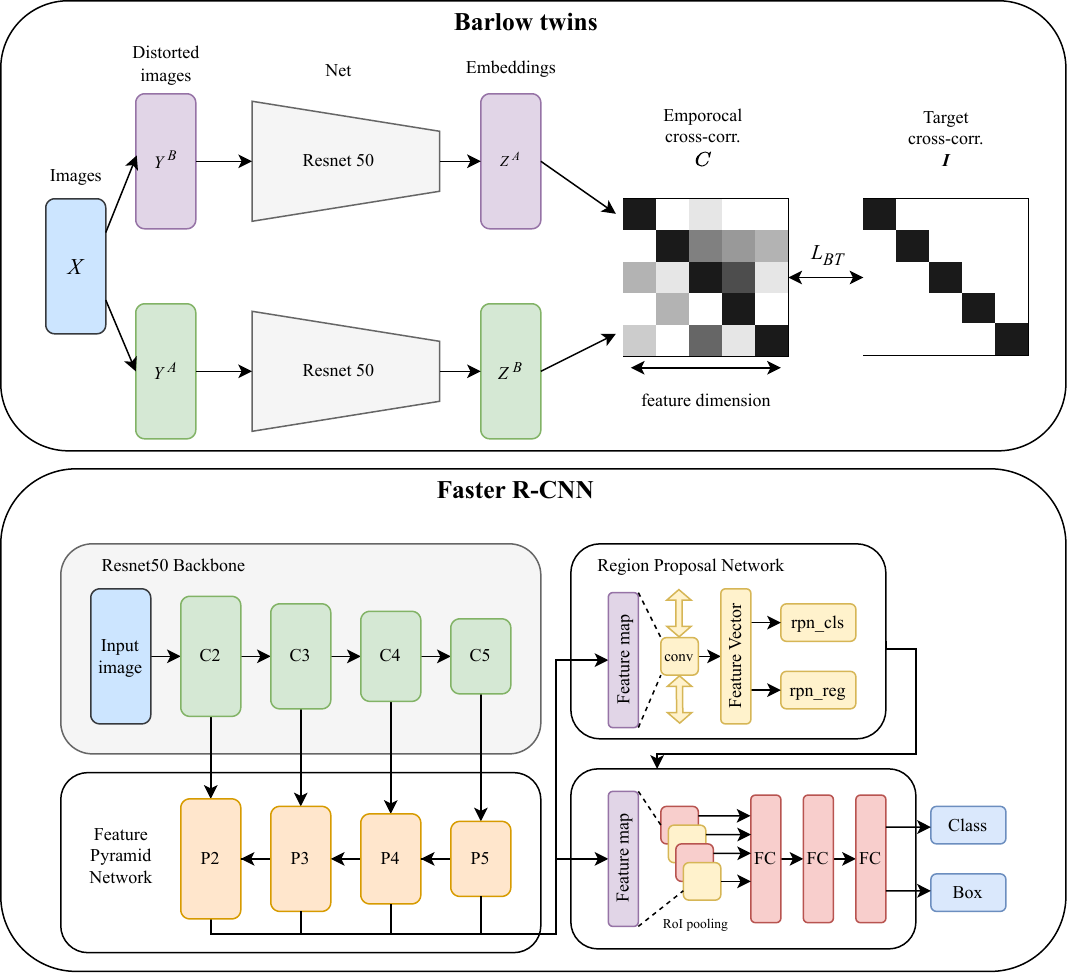}
\caption{Schematic Overview of the Dual-phase Training Framework. The upper panel illustrates the Barlow Twins method in Phase One, where pairs of distorted images are processed through a shared ResNet50 network to produce embeddings. These are then compared using an empirical cross-correlation matrix \textit{C}, striving for the identity matrix \textit{I} to minimize redundancy in feature dimensions, and optimizing the loss function $L_{BT}$. In Phase Two (lower panel), the pre-trained ResNet50 backbone from Phase One is integrated into a Faster R-CNN architecture. It starts with multi-scale feature extraction through the Feature Pyramid Network (FPN), followed by the Region Proposal Network (RPN) that generates object region proposals. The features are then pooled and processed by fully connected (FC) layers to output the final class labels and bounding box coordinates for object detection tasks. }\label{fig3}
\end{figure}

Our training encompasses two primary phases: self-supervised pre-training and subsequent supervised fine-tuning, as shown in Figure \ref{fig3}. Initially, we commenced with a Resnet50 model pre-trained on ImageNet.

In the self-supervised pre-training phase, we applied the Barlow Twins method to the NIH-CXR Dataset to exploit its ability to enhance feature independence, thereby significantly reducing feature redundancy in medical imaging. This approach refined the ImageNet pre-trained model by updating its backbone weights. 

Subsequently, in the supervised fine-tuning phase, we utilize this refined backbone within a Faster R-CNN framework, chosen for its efficacy in precise localization and classification within complex images, to the VinDr-CXR dataset. This step aims to improve the model's task-specific performance further, explicitly enhancing its capabilities in localized anomalies in CXR images.


\subsubsection{Self-Supervised Pre-training}\label{subsubsec3}

We chose the Barlow Twins method for the first stage of our training to ensure that visual features in medical images remain independent from each other, a crucial consideration for improving the performance of medical imaging analysis. 
As outlined by Zbontar et al. \cite{ref8}, 
this approach represents a shift from conventional contrastive learning, introducing a self-supervised learning framework primarily focused on diminishing redundancy. The Barlow Twins method operates on a straightforward yet potent principle: it learns distinctive features by reducing the representational differences between two differently distorted images from the same source as processed by the network. This strategy is instrumental in enabling the model to identify unique and rich features in each image while concurrently minimizing the overlap in features. The process involves generating two distinct variants of an image through data augmentation, followed by their simultaneous processing via two deep neural networks that share identical weights. The objective is to align the network's weights to enhance the similarity in the high-level representations of these image pairs yet ensure that the individual features remain distinct and independent.

The Barlow Twins method might be particularly useful for medical imaging because it extracts features by minimizing the redundancy between representations of perturbed images. In CXR imaging, subtle differences might indicate important health information, and the Barlow Twins can effectively capture these subtle but clinically important features. In contrast to other contrastive learning algorithms like MoCo \cite{ref26} and SimCLR, which construct similarity matrices at the batch dimension, Barlow Twins works at the feature dimension. It aims to assign an independent meaning to each feature dimension. This could lead to a richer feature representation, potentially better adapted to variations in CXR images (e.g., different imaging conditions and pathological states). Moreover, compared to self-supervised learning methods requiring negative samples or complex contrastive mechanisms like SimCLR, Barlow Twins offers a more straightforward training framework, which is particularly important in situations with limited computational resources.

We chose to apply Barlow Twins pre-training on the ImageNet pretrained ResNet50model. Since the ImageNet pre-trained model weights can be easily obtained from the Torchvision library, this step brings no additional cost. We used images from the training set portion of the NIH-CXR dataset for this training phase, with the input image size set to 224*224 pixels. The training was executed on an NVIDIA A100 80G GPU, setting the batch size to 768 to maximize the utilization of this graphics card's capabilities over 600 epochs.

\subsubsection{Fine-tuning Phase}\label{subsubsec3}

In our fine-tuning/transfer learning stage, we utilized the Faster R-CNN \cite{ref27} with Feature Pyramid Network (FPN) \cite{ref28} as our object detector and trained it on the VinDr-CXR dataset. Faster R-CNN, a widely-used object detection framework, comprises two main components: the Region Proposal Network (RPN) \cite{ref28} and the Fast R-CNN detector. First, RPN generates candidate regions for objects, and then the Fast R-CNN detector employs these regions to detect and classify targets. This architecture renders Faster R-CNN particularly efficient in processing complex images. The Feature Pyramid Network (FPN), an architecture frequently employed in object detection, particularly enhances performance with multi-scale targets. It integrates high-level semantic information from deeper layers with detailed information from shallower layers, producing feature maps of varied scales that effectively detect differently sized targets.

We employed the MMdetection \cite{ref29} machine learning toolbox as the platform for Faster R-CNN, utilizing a number of classical image augmentation techniques and maintaining consistent hyperparameters across all experiments. Two different input sizes, 224*224 pixels and 640*640 pixels, were chosen to assess the impact of image size on the model's performance with the pre-trained models. In addition, for comparison, we also conducted experiments using ImageNet pre-trained weights directly.

We implemented a linear evaluation protocol \cite{ref30}\cite{ref31} on the NIH-CXR dataset to comprehensively evaluate the self-supervised learning model's performance in medical imaging. This method examines the model’s feature transfer capability - its ability to adapt learned representations to new tasks. We first resplit the test set of the NIH dataset into two parts: 80\% as an evaluation training set for training a linear classifier and the remaining 20\% as an evaluation test set for assessing model performance.

We adopted two distinct strategies during the evaluation: freezing the backbone weights or fine-tuning the weights. In the freezing backbone strategy, we kept the parameters of the backbone network (i.e., the feature extraction layers) obtained from self-supervised pretraining unchanged. We updated only the weights of the final linear layer. Conversely, under the fine-tuning strategy, we updated parameters across the entire network, encompassing both the self-supervised trained feature extraction layers and the newly added linear classifier layer. We used 100\%, 10\%, and 1\% of the evaluation training set data for training the linear classifier, allowing us to assess the model's performance across different scales of training data.

When evaluating the representation transfer ability of a self-supervised learning model, it is necessary to ensure that the ratio of individual labels in the training and test sets is consistent. We used the Iterative stratification for the multi-label data method \cite{ref32}\cite{ref33} to ensure that the proportions of each label in the evaluation training and test sets were roughly similar. This helped prevent biases due to uneven label distribution, making our evaluation results more reliable and convincing.

\subsection{Results Analysis Process}\label{subsec3}

For the analysis of results, we employed the mean Average Precision (mAP) at an Intersection over Union (IoU) of 50\% as the benchmark for evaluating the performance of our object detection models. mAP is a widely recognized and effective metric in object detection, calculated by averaging precision scores across various object detection confidence thresholds. Specifically, mAP is the mean of the average precision scores for each class. The proportion of correct predictions relative to all predictions for a specific class across different detection confidence thresholds determines the precision score. In the context of CXR abnormality localization, utilizing mAP at an IoU of 50\% is beneficial for capturing clinically significant lesion detections while allowing for a reasonable degree of positional deviation, which is practical for actual clinical applications.

Moreover, we utilized the Area Under the Curve (AUC) as a metric for the linear evaluation protocol. AUC, a standard metric in medical image analysis, balances precision and recall, making it an especially appropriate performance indicator for this field. The AUC metric represents the area under the Receiver Operating Characteristic (ROC) curve, accounting for the model's True Positive Rate (TPR) and False Positive Rate (FPR) at various thresholds. This assessment method balances the model's sensitivity and specificity, enhancing detection rates while controlling false positives. Medical image analysis often deals with imbalanced data, and AUC is robust for imbalanced datasets as it does not rely directly on classification thresholds.

Beyond using mAP and AUC for quantitative analysis, our study also utilized the Ablation CAM (Class Activation Mapping) method to create heat maps for qualitative evaluation. Ablation CAM systematically abates features in the model's final convolutional layer and observes the impact on the output class scores. This process reveals the most influential regions for the model's decision-making. The resulting heat maps delineate areas of interest in CXR images, providing intuitive visual evidence of how our BarlowTwins-CXR model focuses on and recognizes abnormalities.

\section{Results}\label{sec4}

\subsection{Transfer Learning on VinDr Abnormality Localization }\label{subsec4}
\begin{figure}[h]
\centering
\includegraphics[width=\textwidth]{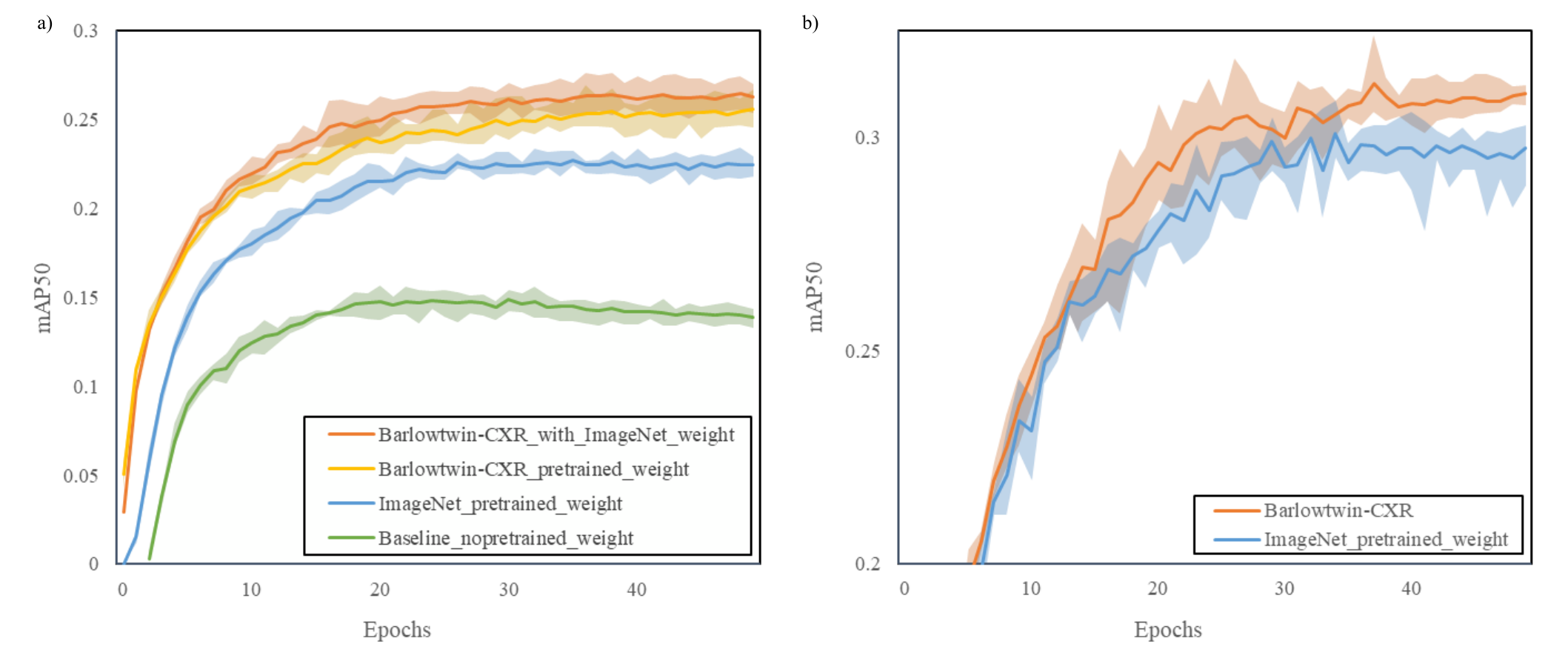}
\caption{Evolution of mAP50 across epochs for different ResNet50 backbones on the VinDr-CXR dataset at 224*224(left) and 640*640(right) resolution. The darker lines represent the average mAP50 of four(left) and five(right) trials with different random seeds, with shaded areas indicating the range between the lowest and highest value. }\label{fig4}
\end{figure}

In this experiment, we examined the efficacy of the ResNet backbone pre-trained by the Barlow Twins-CXR method for abnormality localization on the VinDr-CXR dataset, using two different input resolutions. Consistent hyperparameter settings were maintained across all experiments, ensuring that the performance changes were attributable only to the merits of the pretraining method itself. We visualized the performance of different models such as Barlow twins-CXR pre-training and ImageNet pre-training on the validation set in Figure \ref{fig4}, and tabulated the corresponding mAP performance in Table \ref{Table1}. As depicted in the figure, the baseline model with an untrained ResNet50 backbone reached a final mAP50 score of 0.1342 (95\% CI 0.1306,0.1378), setting a performance baseline without pre-training benefits.

\begin{table}[h]
\caption{mAP50 scores in validation and test sets for models with varying pre-training methods at different input resolutions. }\label{Table1}
\begin{tabular}{@{}llll@{}}
\toprule
BackBone\_weight & Input\_size & mAP50 (val set) & mAP50 (test set) \\
\midrule
baseline\_nopretrained & 224 & 0.1388 (0.1352,0.1424) & 0.1342 (0.1306,0.1378) \\
ImageNet\_pretrained & & 0.2245 (0.2204,0.2286) & 0.2210 (0.2194,0.2226) \\
Barlow\_twins & & 0.2555 (0.2485,0.2626) & 0.2448 (0.2414,0.2482) \\
Barlow\_twins\_from\_ImageNet & & 0.2625 (0.2568,0.2682) & 0.2502 (0.2476,0.2528) \\
\midrule
ImageNet\_pretrained & 640 & 0.2973 (0.2913,0.3033) & 0.280 (0.2757,0.2848) \\
Barlow\_twins\_from\_ImageNet & & 0.3102 (0.3080,0.3125) & 0.289 (0.2826,0.2954) \\
\bottomrule
\end{tabular}
\footnotetext[1]{Scores are presented with 95\% confidence intervals.}
\end{table}

A significant advancement was observed with the ImageNet pre-trained ResNet50, which attained a mAP50 of 0.2210 (95\% CI 0.2194,0.2226), underscoring the value of pre-training in feature representation across disparate image domains.

More strikingly, incorporating the Barlow Twins-CXR strategy led to a rapid performance ascent, achieving a mAP50 of 0.2448 (95\% CI 0.2414 0.2482). It marked an expedited training trajectory and a significant increase in detection performance.

When further enhanced by pre-training from ImageNet, the Barlow Twins-CXR approach yielded the best performance, recording a mAP of 0.2502 (95\% CI 0.2476 0.2528), evidencing the synergetic effect of combining pre-training methodologies.

\begin{figure}[h]
\centering
\includegraphics[width=\textwidth]{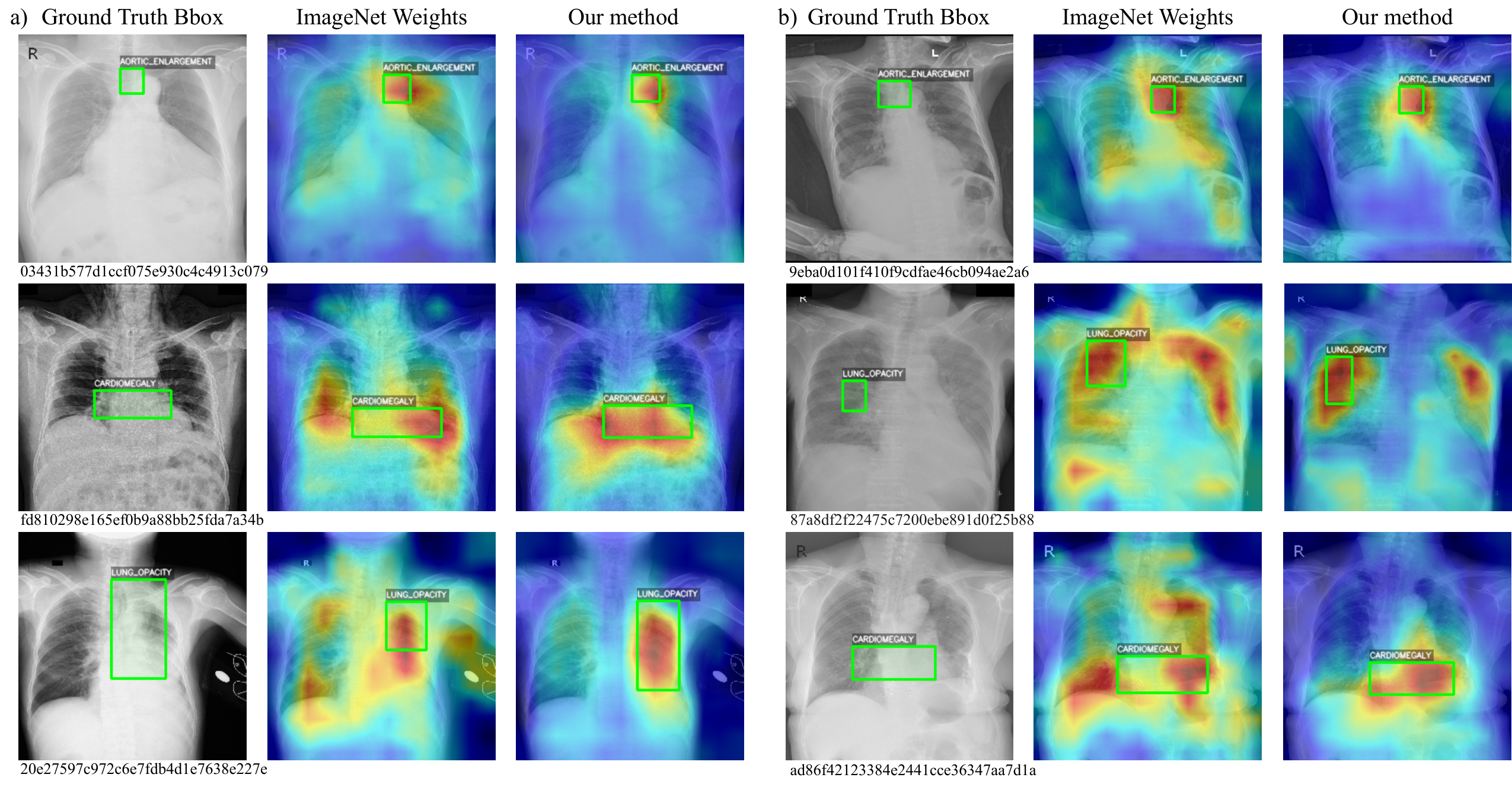}
\caption{Heatmaps were generated from the initial images of the training set(left) and test set(right), indicating successful Bbox predictions by the BarlowTwins-CXR model. Each heatmap corresponds to one accurately predicted bbox, despite multiple bboxes present in each CXR image. Serial numbers below the heatmaps refer to the image numbers in the dataset. }\label{fig5}
\end{figure}

The heat maps generated from the study present a compelling visualization of the performance of the BarlowTwins-CXR method compared to the traditional ImageNet weights approach. We generated heat maps of the first few CXR images of the training and test sets in Figure \ref{fig5}. In each image, our method's heat maps show a more focused alignment with the actual lesion areas marked by the Ground Truth Bbox. This indicates a higher precision in localizing and identifying pathological features with BarlowTwins-CXR, potentially offering more targeted information for clinical diagnoses. Notably, in cases of cardiomegaly and lung opacity, the concentration and localization of the heatmaps from BarlowTwins-CXR are visibly superior to those derived from ImageNet weights, further affirming the efficacy of our approach in enhancing CXR image analysis.

Upon escalating the input resolution to 640 * 640 pixels, both ImageNet and Barlow Twin-CXR weighted models saw performance improvements due to the increased detail in the CXR images. Nonetheless, the performance differential between the two narrowed, indicating that the higher resolution somewhat mitigates the distinct advantages of self-supervised pre-training.

This points to intriguing future research avenues, such as refining image resolution parameters during pre-training and fine-tuning phases and investigating whether higher-resolution pre-training could elevate model performance. It also accentuates the necessity of tailoring deep learning model design to specific tasks, considering factors like image resolution and feature granularity.

Overall, implementing the Barlow Twins-CXR method on the VinDr dataset resulted in substantial gains despite its data limitations and the inherent challenges of CXR abnormality localization. An 11.5\% performance enhancement over the baseline and a 2.8\% increment over ImageNet pre-trained models were observed on the mAP50 metric. Such marked improvements confirm the Barlow Twins-CXR strategy's prowess in addressing domain inconsistencies, thereby fine-tuning naturally derived image weights for better applicability in CXR image analysis and beyond in medical imaging.

\subsection{Linear Evaluation Protocol  }\label{subsec4}
\begin{figure}[h]
\centering
\includegraphics[width=0.5\textwidth]{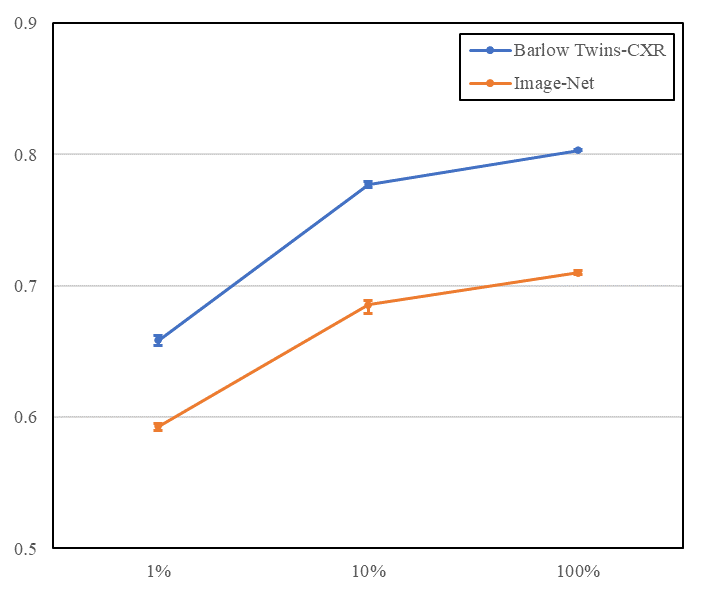}
\caption{AUC Scores with Error Bars for NIH-CXR Classification - This figure displays the AUC scores of linear models with Barlow Twins-CXR versus ImageNet weights across various dataset sizes (1\%, 10\%, 100\%). As indicated by higher AUC scores, models using Barlow Twins-CXR consistently outperform those with ImageNet pre-training. Error bars represent the range of scores across five experiments. }\label{fig6}
\end{figure}

In this experiment, we evaluated the impact of Barlow Twins-CXR pre-training versus traditional ImageNet pre-training on the linear classification performance within the NIH-CXR dataset. We adhered to the linear evaluation protocol, freezing the backbone of the linear classifier and updating only the final linear layer's weights. This approach was applied across training datasets of varying sizes - 1\%, 10\%, and 100\%, results of these experiments are presented in Figure \ref{fig6} and Table \ref{tabel2}.

\begin{table}[h]
\caption{AUC scores in validation and test sets for of linear models with varying pre-training methods at 224 and 640 input resolutions. }\label{tabel2}
\begin{tabular}{@{}lccc@{}}
\toprule
Model & 1\% & 10\% & 100\% \\
\midrule
Barlowtwin-CXR & 0.6586 (0.6556, 0.6616) & 0.7773 (0.7756, 0.7790) & 0.8031 (0.8027, 0.8035) \\
Image-Net & 0.5932 (0.5913, 0.5951) & 0.6855 (0.6822, 0.6889) & 0.7098 (0.7089, 0.7107) \\
\bottomrule
\end{tabular}
\footnotetext[1]{Scores are presented with 95\% confidence intervals.}
\end{table}

The results show that at a training data size of 1\%, the Barlow Twins-CXR pre-trained model demonstrated a significant advantage, achieving an AUC of 0.6586 (95\% CI 0.6556,0.6616) compared to 0.5932 (95\% CI 0.5913,0.5951) for the ImageNet pre-trained model. As the training data size increased to 10\% and 100\%, the AUCs for the Barlow Twins-CXR pre-trained model reached 0.7773 (95\% CI 0.7756,0.7790) and 0.8031 (95\% CI 0.8027,0.8035), respectively, while the ImageNet pre-trained model scored 0.6855 (95\% CI 0.6822,0.6889) and 0.7098 (95\% CI 0.7089,0.7107).

Notably, the incremental gains for both pre-training methods diminished with larger data sizes, suggesting that the performance boost provided by additional data becomes marginal when only the linear layer is updated.

These findings highlight the Barlow Twins-CXR pre-training method's superiority over ImageNet pre-training across various dataset sizes, especially in data-limited scenarios. This demonstrates the promise of self-supervised learning in enhancing medical image analysis, particularly when annotated data is scarce.

\subsection{End-to-End Finetuning  }\label{subsec4}

\begin{figure}[h]
\centering
\includegraphics[width=0.5\textwidth]{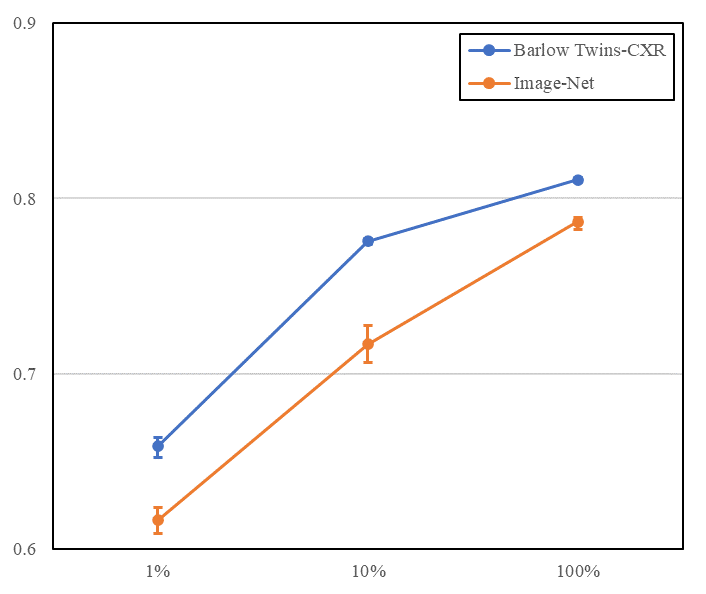}
\caption{AUC Scores with Error Bars for NIH-CXR Classification - This figure displays the AUC scores of models fine-tuned end-to-end with Barlow Twins-CXR versus ImageNet weights across various dataset sizes (1\%, 10\%, 100\%). Higher AUC scores indicate that models using Barlow Twins-CXR consistently outperform those with ImageNet pre-training. Error bars represent the range of scores across five experiments. }\label{fig7}
\end{figure}

In our end-to-end experiments, where we permitted updates to all model layers, the Barlow Twins-CXR pre-trained ResNet50 backbone consistently outperformed the ImageNet pre-trained equivalent across all training set sizes. The results of these experiments are presented in Figure \ref{fig7} and Table \ref{tabel3}.

\begin{table}[h]
\caption{AUC scores in validation and test sets for models fine-tuned end-to-end with varying pre-training methods at 224 and 640 input resolutions. }\label{tabel3}
\begin{tabular}{@{}lccc@{}}
\toprule
Model & 1\% & 10\% & 100\% \\
\midrule
Barlowtwin-CXR & 0.6585 (0.6544, 0.6627)
 & 0.7756 (0.7745, 0.7768) & 0.8107 (0.8098, 0.8116) \\
Image-Net & 0.6163 (0.6110, 0.6216) & 0.7168 (0.7093, 0.7243) & 0.7866 (0.7843, 0.7889) \\
\bottomrule
\end{tabular}
\footnotetext[1]{Scores are presented with 95\% confidence intervals.}
\end{table}

At a 1\% training data size, the Barlow Twins-CXR model achieved a 4.2\% higher AUC than the ImageNet counterpart.

With 10\% and 100\% data sizes, the Barlow Twins-CXR model maintained leads of approximately 5.9\% and 2.5\%, respectively. Notably, the magnitude of improvement over the frozen backbone setup was less marked, suggesting that the wealth of features learned during self-supervised training reduces the margin for additional gains during subsequent fine-tuning.

Overall, these end-to-end fine-tuning results suggest that comprehensive learning across all model layers may elevate the risk of overfitting, particularly when data is scarce. The narrowing performance differential between the two pre-training strategies with increasing data volume indicates that the distinction between domain-specific (Barlow Twins-CXR) and generalized (ImageNet) pre-training becomes less substantial with larger datasets. This trend implies that the influence of the pre-training strategy on the final performance of models may diminish as the size of the medical image dataset grows.

\section{Discussion}\label{sec5}
Our study demonstrates that the BarlowTwins-CXR approach effectively utilizes unannotated CXR images for learning valuable representations and enhances transfer learning efficiency from ImageNet, thus addressing issues of domain inconsistency. This leads to quicker training and improved performance on tasks like abnormality detection in the VinDr-CXR dataset. Barlow Twins-CXR excels across various input resolutions, outshining models pre-trained on ImageNet.

One of the primary limitations of our study is the scarcity of CXR datasets with bounding box. Our reliance on public datasets, due to the absence of a private dataset, may limit the generalizability of our findings. Additionally, the computational cost of the BarlowTwins pre-training remains substantial. For a dataset size of 112,120 images with an image size of 224*224 pixels, the training process required two days on an NVIDIA A100 80G GPU. This significant resource requirement constrained our ability to experiment with higher image resolutions, which could potentially enhance the model's performance.

\section{Future Work}\label{sec6}
Our future endeavours include developing a demo interactive system for deployment and testing in emergency rooms. It will allow practical evaluation of the model's effectiveness in a clinical setting and facilitate the collection of a proprietary dataset. Additionally, we plan to explore more advanced self-supervised learning methods, object detection frameworks, and backbone networks to refine our approach further. The continuous evolution of these technologies promises to address some of the current limitations and expand the applicability and accuracy of our model in medical image analysis.

\section{Conclusions}\label{sec5}

The results of this study provide strong support for the application of self-supervised learning in the field of abnormality detection, especially valuable in environments where radiologists face high workloads but the corresponding data labelling resources are scarce. A critical aspect of this approach is its adaptability to regional variations in CXR image, attributable to differences in imaging equipment, patient demographics, and other locale-specific factors \cite{ref34}\cite{ref35}. Such variations often impede the cross-regional applicability of a model, thus limiting its generalizability. By employing the BarlowTwins-CXR strategy, research organizations can transfer pre-trained backbone networks to local datasets tailored to the unique characteristics of their regional data. 

Our findings might also have significant implications for clinical practice, suggesting that this strategy could be a game-changer in aiding radiologists to interpret CXR images efficiently. This technology promises to reduce diagnostic times, potentially increasing patients' throughput and improving the overall quality of care. Given its capacity for fine-tuning to specific regional characteristics, our approach holds particular promise in areas where standardization of medical imaging presents challenges.

In summary, the BarlowTwins-CXR approach demonstrates the potential of AI to enhance healthcare delivery. By integrating cutting-edge technology with clinical needs, we aim to pave the way for innovative solutions that benefit healthcare professionals and patients.

\section{Abbreviations}\label{sec6}
AP: anterior-posterior\\
AUC: area under the receiver operating characteristic curve\\
CAM: Class Activation Mapping\\
CIUSSS: Centre intégré universitaire de santé et de services sociaux\\
CXR: chest X-ray radiography\\
FC: Fully connected layer\\
FPN: Feature Pyramid Network\\
FPR: False Positive Rate\\
IoU: Intersection over Union\\
ROC: receiver operating characteristic\\
ROI: region of interest\\
mAP: mean Average Precision\\
PA: posterior-anterior\\
TPR: True Positive Rate\\
WBF: Weighted Box Fusion\\
YOLO: You Only Look Once

\section{Declarations}\label{sec7}
\subsection{Ethics approval and consent to participate  }\label{subsec7}
All methods were performed under relevant guidelines and regulations (e.g., Declarations of Helsinki). The studies reported in this manuscript used reputable public datasets and did not require any additional data involving human participants, human data, or human tissue.
\subsection{Consent for publication  }\label{subsec7}
Not applicable

\subsection{Availability of data and materials  }\label{subsec7}
The datasets generated and/or analysed during the current study are available in the VinDr-CXR \cite{ref10} and NIH-CXR\cite{ref24} repository:
\href{https://physionet.org/content/vindr-cxr/1.0.0/}{VIndr-CXR} and \href{https://nihcc.app.box.com/v/ChestXray-NIHCC}{NIH-CXR}.

\subsection{Competing interests  }\label{subsec7}
The authors declare that they have no competing interests

\subsection{Funding  }\label{subsec7}
No external funding was associated with this research study.

\subsection{Authors' contributions  }\label{subsec7}
HS designed the research methodology, analyzed data, was responsible for experiments and results visualization, and participated in manuscript drafting and revision.
LM assisted in developing the research methodology and contributed to the drafting and revision of the manuscript.
JFS collected and interpreted data, and provided expertise in statistical analysis.
DL contributed to the study design, offered statistical analysis expertise, assisted in interpreting results, and played a significant role in the critical revision of the manuscript.

All authors read and approved the final manuscript.

\subsection{Acknowledgements  }\label{subsec7}
The authors wish to express their gratitude to CIUSSS du centre-sud-de-l'île-de-montréal for the computational resources and support provided, which were essential for the research conducted as part of the graduate internship program. We are especially thankful to our department director, Mathieu Mailhot, for his mentorship and to Chen Cheng for his collaborative efforts and valuable contributions to this project. Their expertise and insights have been greatly appreciated and substantially enhanced this work's quality.







\bibliography{sn-bibliography}

\end{document}